\documentclass{article}
\usepackage{lipsum}
\usepackage{mwe}
\usepackage[utf8]{inputenc}
\usepackage{longtable}
\usepackage{blindtext,alltt}
\usepackage{xcolor}
\usepackage{float}
\usepackage{geometry}
\usepackage{csquotes}
\usepackage[utf8]{inputenc}
\usepackage[english]{babel}
\usepackage[
    backend=biber,
    style=numeric,
    citestyle=numeric,
]{biblatex}
\usepackage{titlesec}
\titleformat{\paragraph}
{\normalfont\normalsize\bfseries}{\theparagraph}{1em}{}
\titlespacing*{\paragraph}
{0pt}{3.25ex plus 1ex minus .2ex}{1.5ex plus .2ex}

\title{Catastrophic Forgetting in the Context of Model Updates}
\author{Hillary Sanders \& Rich Harang}

\begin{document}
\maketitle

{
\color{red}
}

\begin{abstract}
A large obstacle to deploying deep learning models in practice is the process of updating models post-deployment (ideally, frequently). Deep neural networks can cost many thousands of dollars to train. When new data comes in the pipeline, you can train a new model from scratch (randomly initialized weights) on all existing data. Instead, you can take an existing model and fine-tune (continue to train) it on new data. The former is costly and slow. The latter is cheap and fast, but catastrophic forgetting generally causes the new model to 'forget' how to classify older data well. There are a plethora of complicated techniques to keep models from forgetting their past learnings. Arguably the most basic is to mix in a small amount of past data into the new data during fine-tuning: also known as 'data rehearsal'. In this paper, we compare various methods of limiting catastrophic forgetting and conclude that if you can maintain access to a portion of your past data (or tasks), data rehearsal is ideal in terms of overall accuracy across all time periods, and performs even better when combined with methods like Elastic Weight Consolidation (EWC). Especially when the amount of past data (past 'tasks') is large compared to new data, the cost of updating an existing model is far cheaper and faster than training a new model from scratch.

\end{abstract}
 
\section{Introduction}
Catastrophic Forgetting - the tendency of deep neural networks to 'forget' previously learned information upon learning new information - has been studied since 1989\cite{catference}. This is most clearly demonstrated when models are given separate tasks to learn sequentially, but the effect is also at play whenever a model is learning any type of information sequentially, where that information changes in distribution over time. Real-world applications of machine learning very often have new (training) data coming in over time. In order to release new models trained on this new information, machine learning developers can retrain an entire model from scratch (randomly initialized weights) using all existing training data, but this is computationally very costly. Another option is to take an existing model trained on past data, and simply fine-tune it on the new data that has come in. But new data is coming from (generally) a slightly different 'distribution' than old data, and so especially when this change in distribution is large, the catastrophic forgetting effect becomes clear upon fine-tuning.
\newline
\newline
In the malware detection space, the distribution of malicious and benign content is always changing. New malware is always being produced, as well as new benign content. As a result, a model trained on data up until time $t$ will generally perform extremely well on a holdout set from before $t$ (coming from the same parent distribution as the training data), but it's accuracy will, on average, decay validated against $t+n$ holdout sets with increasing $n$.
\newline
\newline
\begin{figure}[!ht]
    \centering
        \includegraphics[width=.75\textwidth]{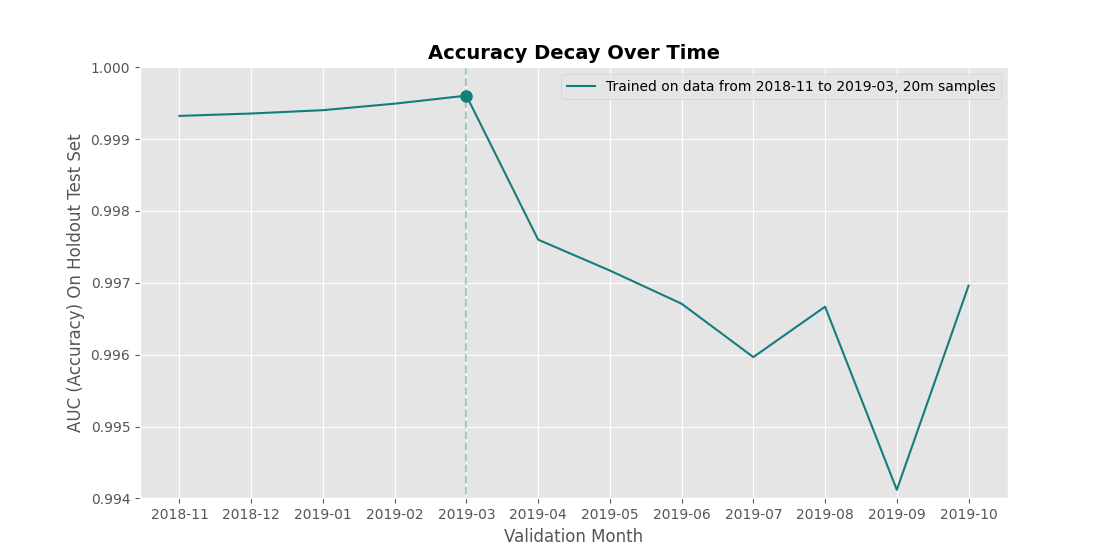}
        \caption{Model performance on holdout validation data over time (timestamp value is based on the first-seen time of a particular sample). Model was trained on training data up until 2019-03. While performance on holdout validation data is high in the months overlapping with training data, the model's accuracy quickly decays on holdout validation data coming from after 2019-03.}
\end{figure}

Simply fine-tuning existing models with new data, however, results in models that perform well on data similar to the fine-tuned data, but perform poorly on past data (which we still want our models to be able to classify well). Figure 2 contrasts a model that has been sequentially fine-tuned on monthly data (colored lines) vs a model that has been fully retrained from scratch. Note how the red line (sequentially fine-tuned) does more poorly on past months, than the black line (retrained from scratch).
\newline
\newline

\begin{figure}[!ht]
    \centering
        \includegraphics[width=.75\textwidth]{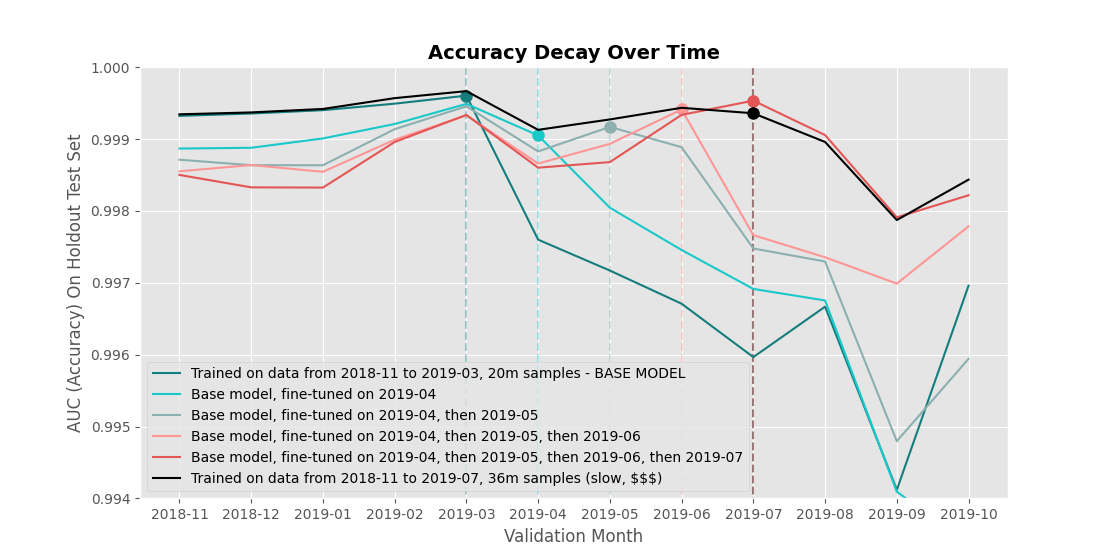}
        \caption{Here, a base model trained on data up until 2019-03 (the dark teal line) is sequentially fine-tuned on data from the following months: 2019-04 (light teal), then 2019-05 (grey), then 2019-06 (light pink), and finally 2019-07 (the red line). The black line represents a model that was retrained from scratch (with randomly initialized weights) on all data up until 2019-07. While the black line is far more computationally costly to produce each month, note how it performs better on older data than its sequentially fine-tuned comparison model, the red line.
        }
\end{figure}

 If you can get sequentially fine-tuned models (red line) to match or beat performance of a model trained all at once (black line) by minimizing catastrophic forgetting, then you have a path to deploying accurate models very quickly and very cheaply. In this paper, we compare various methods of minimizing this catastrophic forgetting effect. Arguably the most basic approach, one that was suggested in the 90s by X [cite], is to simply mix in data associated with past 'tasks' (older data) into the fine-tuning dataset. We also tested approaches that don't require access to past data: model averaging, L2 regularization on weight movement, and Elastic Weight Consolidation (EWC) regularization \cite{ewc}. Our results showed that by far the most effective approach is data rehearsal.

\section{Related Work}
Catastrophic Forgetting was introduced as an issue in 1989\cite{catference} by McCloskey and Cohen. In 1995 Robins \cite{rehearsal} discussed a model training set up where each new item (task) is a single sample, but instead of training a model by sequentially learning on each item individually, items were learned alongside 3 past items during each epoch. This was termed 'sweep rehearsal', and helped improve forgetting. This concept generalizes to situations in which each item (new 'task') is composed of many samples, and data associated with old tasks is mixed in with new task data during model training.
\newline
\newline
Data rehearsal requires access to past data, which isn't always possible. Much work has been done on the subject of limiting catastrophic forgetting when you don't have permanent access to past task data. Atkinson et. al.\cite{psuedorehearsal} (among others) use a psuedo-rehearsal system to generate (via a deep generative neural network) items representative of previous tasks.
{

\section{Experiments}

Our experimental setup was as follows. We experimented with our PE malware detection model, which is deep neural network consisting of five large fully connected layers (sizes 1024, 768, 512, 512, 512), followed by output layers (primarily the `is\_malware' output). We grabbed data and extracted features for 12 separate months (each treated as a 'task' in our set up), consisting of about four million training samples each (about a million leftover for validation and testing). The first five months were used to train an initial 'base' model. To test a method, the base model was fine-tuned on data from the sixth month (using the chosen method, e.g. L2 regularization, EWC, etc), the resulting model was trained on the seventh month, that resulting model was trained on the eight month, and that resulting model was trained on the ninth month. The final three months were kept a time-split validation holdout. We chose to apply the methods sequentially on four separate months in order to better simulate real-world model updates (if a method cannot be applied iteratively and still work, it is of \textit{much} less use). Unless otherwise stated, all models were trained for ten epochs during each training or fine-tuning session. As a comparison, we also trained a model on the first nine months of data all at once. Although theoretically a sequentially updated model could perform better, the basic idea is that a sequentially updated model with no catastrophic forgetting will perform the same as a model that was trained all at once (the latter being far more cost-prohibitive to train for each new model update).
\newline
\newline
How to best evaluate a model is dependant on what the model is to be used for. We chose to use average AUC on holdout test sets across all twelve months as our main metric (using the final sequentially updated model that has seen the first nine months of data) . We also separately show (see Table 1): average AUC across the first five `base' months, the four `update' months, and the final three `future' months.

\subsection{Model Averaging}
Simply training multiple models per time period and averaging the results seems like a potentially reasonable approach. It would result in a much larger and harder to deploy ensemble model, but each additional model would be fast to train. Our model averaging results, however, were very poor. Averaging the predictions of multiple separate models, each trained on one month's worth of data (for the same number of epochs as the sequentially fine-tuned model, making training costs equal) performed much worse than a sequentially fine-tuned model.
\newline
\newline
\begin{figure}[H]
    \centering
    \includegraphics[width=.75\textwidth]{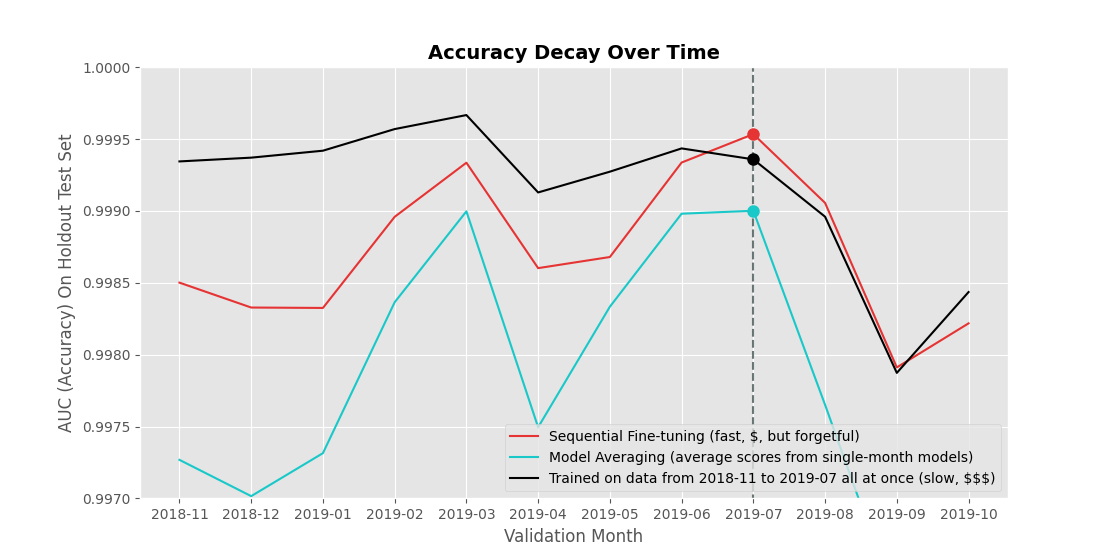}
    \caption{Model averaging results are shown in blue. (Note that the y-limits in the Experiments figures are different than those shown earlier in the paper.)}
    \label{fig:modelavg}
\end{figure}

\subsection{Regularization}
One option to limit `forgetting' is to suppress the movement of parameters during fine-tuning. You can do this naively, like via $L2$ regularization, or you can use more complicated methods like Elastic Weight Consolidation (EWC)\cite{ewc} to penalize the movement of parameters that are estimated to be important to past tasks.
\subsubsection{L2 Regularization}
Figure 3 shows $L2$ regularization: an $L2$ loss penalty is added to each parameter with respect to the current parameter's value $\theta_{c} $ during fine-tuning, and the previous model's parameter value $\theta_{p}$. The further away you move from the previous model's parameter value, the higher the penalty:$(\theta_{c} - \theta_{p})^2 $. Training loss (denoted $ L_{c}^{`}(\theta)$ in the equation below) is simply standard training loss of the current model $L_{c}(\theta)$, plus the $L2$ regularization penalty:
\begin{equation}
    L_{c}^{`}(\theta) = L_{c}(\theta) + \sum_i^{n}{ (\theta_{i, c} - \theta_{i, p})^2 }
\end{equation}

\begin{figure}[H]
    \centering
    \includegraphics[width=.75\textwidth]{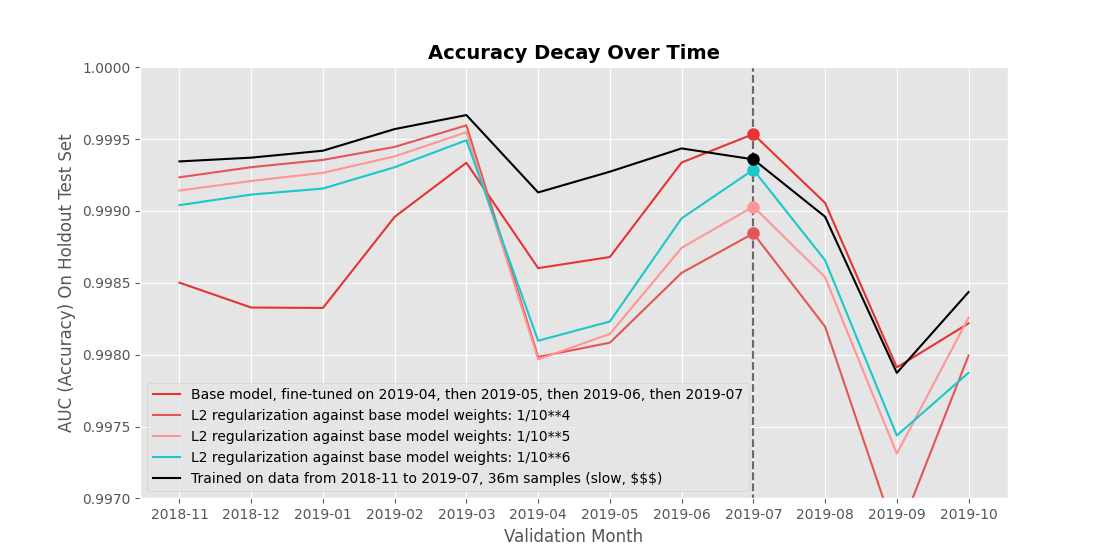}
    \caption{L2 regularization. The colored lines represent the original base model trained up until 2019-03, and then sequentially fine-tuned using different levels of $L2$ regularization.}
    \label{fig:l2}
\end{figure}

As expected, forgetting is reduced but learning is reduced too. There's a very clear trade-off, with no happy medium that attains the performance of the trained-from-scratch model.

\subsubsection{EWC Regularization}
Elastic Weight Consolidation (EWC)\cite{ewc} is a regularization method developed by researchers at Deep Mind in 2017. EWC is similar to $L2$ regularization, but each parameter's movement penalty is scaled by the parameter movement squared ($(\theta_{c} - \theta_{p})^2 $: standard $L2$ regularization), multiplied by EWC's estimate of parameter importance to past tasks (in our case, the ability to classify past data). The idea is to think about your base model as a Bayesian prior of the parameter estimates - then fine tuning is just applying more data to approximate the posterior. Essentially, you apply regularization that simulates the prior distribution. EWC assumes that the prior is normally distributed: with mean given by the base model's parameters, and variance given by the inverse of the Fisher Information matrix diagonal $F$.
Instead of minimizing loss during fine-tuning with respect to your new training data, you minimize an estimate of your loss with respect to both your past $p$ and current $c$ data:

\begin{equation}
    L_{p,c}(\theta)  =  L_{c}(\theta) +  \lambda L_{p}(\theta) = L_{c}(\theta) + \sum_i^{n}{\frac{\lambda}{2} *F_{i, p} * (\theta_{i, c} - \theta_{i, p})^2 }
\end{equation}
\newline
\newline
So what we have here is $L2$ regularization, scaled by a scaling parameter $\frac{\lambda}{2}$ and by this $F_i$ value for each parameter $\theta_i$.
\newline
\newline
Elastic Weight Consolidation (EWC)\cite{ewc} is similar to $L2$ regularization, but each parameter's movement penalty is scaled by the movement squared (standard $L2$ regularization), multiplied by EWC's estimate of parameter importance to classifying past data. EWC estimates this using the diagonal of the Hessian of the negative log likelihood with respect to parameters (where this loss corresponds to accuracy on past task(s)). This requires saving second partial derivatives of your model's loss with respect to its parameters during each training round, but doesn't require saving the data used to train the model each round.
\newline
\newline
    The specifics of the formula appear to come from using Taylor Series / Laplace Approximation. Loss with current parameters on current data (fine-tuning data) is known, but estimating loss with current parameters on past data is not known. Without access to past data during fine-tuning, but \emph{with} access to saved partial derivatives of loss on past data with previous parameters $\theta_{p}$, you can estimate the loss $L_p$ on past data with current parameters $\theta_{c}$ via Taylor Series:
    \begin{equation}
        L_p(\theta_{c}) = L_p(\theta_{p}) +   \frac{\partial L_p}{\partial \theta_p}(\theta_c - \theta_p) + \frac{1}{2} \frac{\partial^2 L_p}{\partial \theta_p^2}(\theta_c - \theta_p)^2 + ... 
    \end{equation}
    
    Because we're minimizing loss through gradient descent, we only care about terms that aren't constant with respect to $\theta_c$. So the first term on the right hand side of of the equation we can drop. The second term we can assume is 0 because $L_p$ was at a local minimum with previous parameters $\theta_p$ (before fine-tuning), and the fourth term ($...$) we're choosing to drop as ``small change". What we really want to estimate is the loss of both current and previous data on current parameters $\theta_c$, so adding $ L_c(\theta_{c})$ to both sides of our equation leads us to:
    \begin{equation}
        L_{p,c}(\theta_c) = L_c(\theta_{c}) +  \frac{1}{2} \frac{\partial^2 L_p}{\partial \theta_p^2}(\theta_c - \theta_p)^2 + constant
    \end{equation}
    Replacing $\frac{1}{2}$ with the scaling term $\frac{\lambda}{2}$, leads us to an equation that looks very similar to the one the EWC paper shows. The formula we arrived at represents fisher information $F$ as $\frac{\partial^2 L_p}{\partial \theta_p^2}$. But is that what Fisher Information is?
    \newline
    \newline
     Well, with your model's (negative log likelihood) loss as $L_p$, the Fisher diagonal is defined as $E(-\frac{\partial L_p}{\partial \theta} * -\frac{\partial L_p}{\partial \theta})=E(\frac{\partial L_p}{\partial \theta} * \frac{\partial L_p}{\partial \theta})$: the negatives coming from the 'negative' log likelihood and then cancelling out.\footnote{It's interesting to note that when loss is at a local minimum and thus $E(\frac{\partial L_p}{\partial \theta})=0$, $E(\frac{\partial L_p}{\partial \theta} * \frac{\partial L_p}{\partial \theta})$ is the same as the $variance(\frac{\partial L_p}{\partial \theta})$}.
    \newline
    \newline
    Under certain regularity constraints (which don't hold particularly true in this case), the Fisher diagonal value \emph{is} equal to the negative expected second partial derivative of the Loss $L$ with respect to $\theta$: $- E(-\frac{\partial^2L}{\partial \theta^2}) = E(\frac{\partial^2L}{\partial \theta^2})$ - i.e., the diagonal of the Hessian of the negative log-likelihood (loss) with respect to parameters: what we had in our derivation formula. If you switch out $E(\frac{\partial^2L}{\partial \theta^2})$ for $E(\frac{\partial L_p}{\partial \theta} * \frac{\partial L_p}{\partial \theta})$, you arrive at the formula Deep Mind proposed.\footnote{the derivation was not published alongside the paper, so this is our guess at how they derived their formula!}
\footnote{A more intuitive way to think about this is that  $variance(\frac{\partial L}{\partial \theta})$ might be a nice way to estimate how sensitive your  model is to changes in $\theta$. When loss is greatly affected by small changes to a given parameter $\theta_i$ (i.e. $variance(\frac{\partial L}{\partial \theta_i})$) is high), then the `confidence' that $\theta_i$ should not be changed is high, so the variance in the prior distribution should be small (tight). The added regularization term simulates this confidence, so that parameters the model is very sensitive to don't end up being changed much.}
    \newline
    \newline
    While the derivation uses the Hessian form of Fisher Information,
    in practice, we found that we had to use the $E(\frac{\partial L}{\partial \theta} * \frac{\partial L}{\partial \theta})$ version to attain any sort of good performance. The reason is that if, for any parameter $i$, your $F_{i,p}$ term is less than zero, then the resulting regularization term will just keep on pushing $\theta_i$ further and further away from $\theta_{i,p}$ (the previous value), which will tend to wreck any model (including ours). The $E(\frac{\partial L}{\partial \theta} * \frac{\partial L}{\partial \theta})$ form of fisher $F$ is guaranteed to be positive, while the $E(\frac{\partial^2L}{\partial \theta^2})$ form, not so much. In practice, we found that about a third of our $E(\frac{\partial^2L}{\partial \theta^2})$ estimates tended to be negative, making that approach unusable.
\begin{figure}[H]
    \centering
    \includegraphics[width=0.75\textwidth]{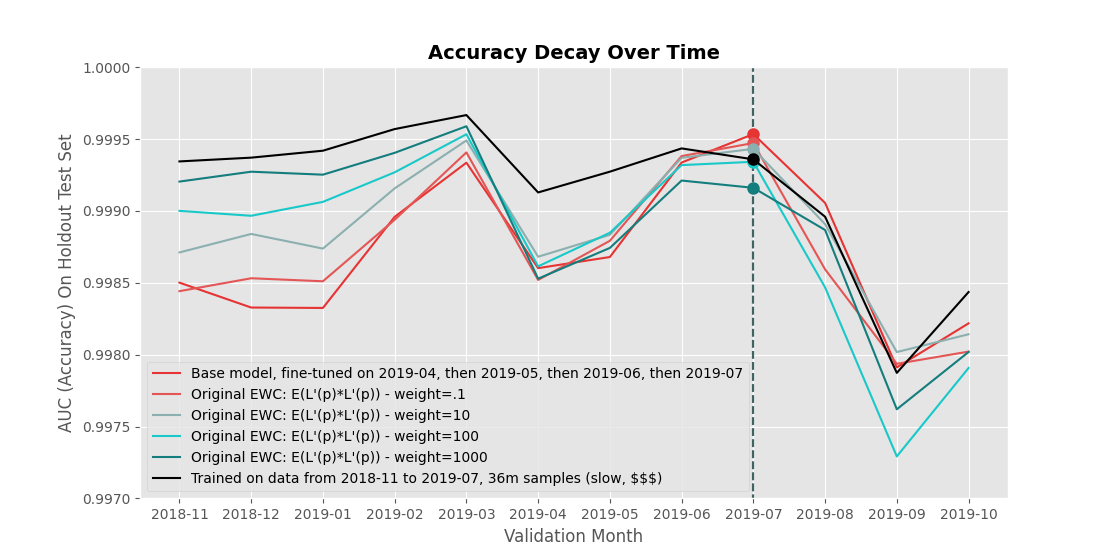}
    \caption{Elastic Weight Consolidation}
    \label{fig:ExampleFigure}
\end{figure}
EWC regularization performed better than basic L2 regularization, but not by an outstanding amount. We theorize that the larger and more complicated a model is, the more error is introduced by the taylor-series estimations and approximations involved in EWC.

\subsection{Data Rehearsal}
Robins \cite{rehearsal} discussed the concept of rehearsal in 1995, shortly after the concept of catastrophic forgetting was introduced. Simply put: you mix in data from past tasks while training on new ones. It works really well, but it does mean you need to maintain access to your older data (or at least an \emph{iid} (independent and identically distributed, i.e. random) subsample of it), which isn't always possible. Mixing in old data also increases the amount of training data you have, so it takes longer to fine-tune a model one epoch.
\newline
\newline
Initially, to make comparisons to other methods equal, we fixed the fine-tuning epoch sizes to 4 million, and tested out various proportions from past data (see Figure 6). For example, a 50\% rehearsal indicates that during fine-tuning, 2 million samples come from the new month's data, and 2 million samples come from past data (sampled uniformly unless otherwise stated) - meaning that the model misses out on much new data each epoch.

\begin{figure}[!ht]
    \centering
    \includegraphics[width=.75\textwidth]{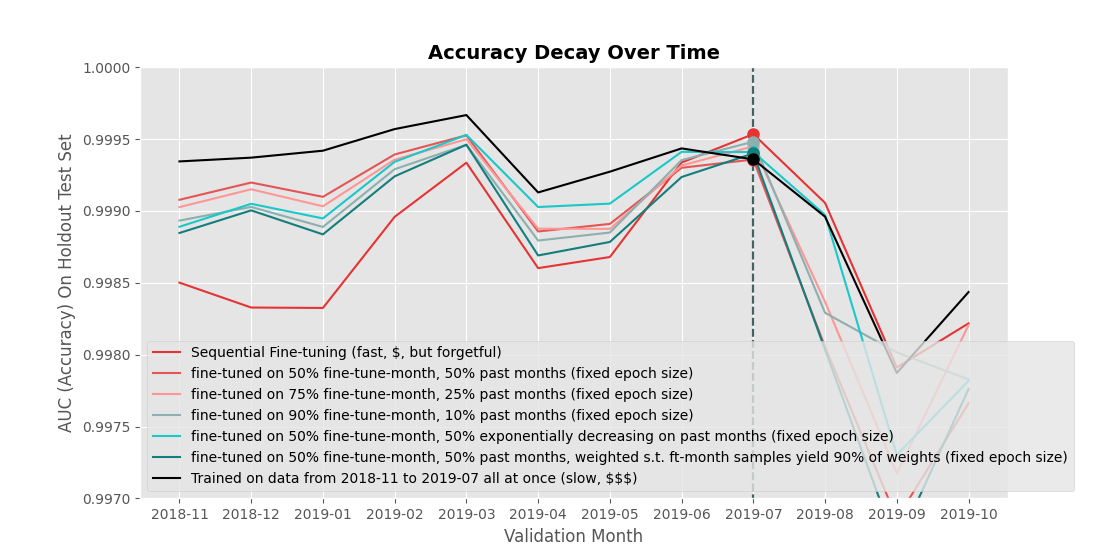}
    \caption{Various forms of data rehearsal with fixed number of training samples each epoch. Each form mixes in varying amounts of past data, but all train for the same number of epochs on the same total number of samples.}
    \label{fig:datarehearsal}
\end{figure}

Not fixing the fine-tuning epoch size (i.e. allow it to increase with added past data) yielded far better results (see Figure 7). However, doing so increases the computational cost of each model update - for example, if you rehearse old data at a rate of 50\%, epoch size jumps to 8 million, doubling the computational cost of training. This is still generally much cheaper computationally than retraining the model from scratch on all available data, but is no longer a computationally fair comparison to the other methods presented above.

\begin{figure}[!ht]
    \centering
    \includegraphics[width=.75\textwidth]{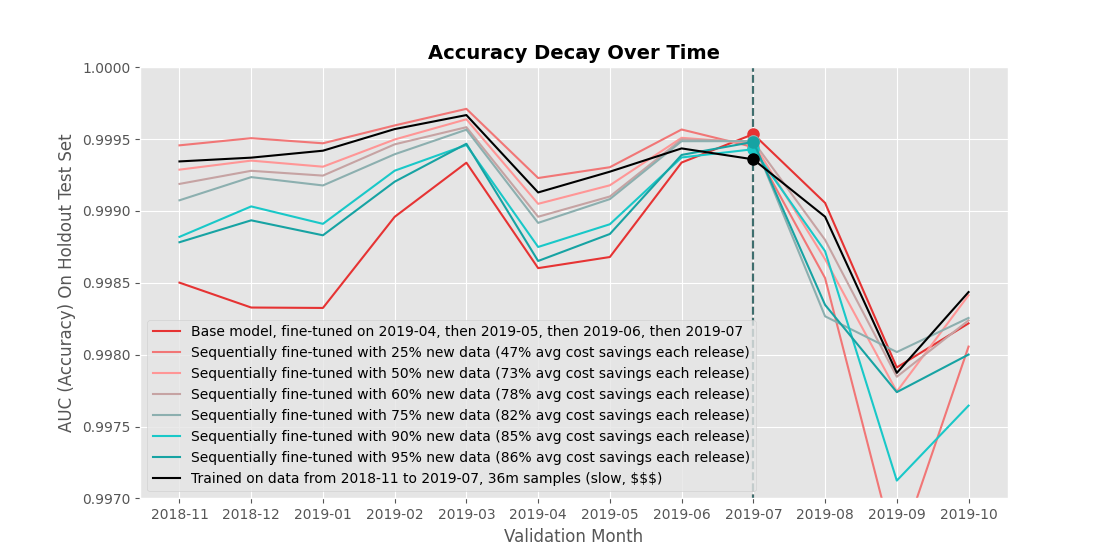}
    \caption{Rehearsal without epoch size limits (i.e. fine-tuning updates are slower amd more costly). 50\% rehearsal (2x time to train, light-pink) achieved the best average AUC.}
    \label{fig:datarehearsal2}
\end{figure}

\subsection{Conclusion}

\begin{figure}[H]
    \includegraphics[width=.65\textwidth]{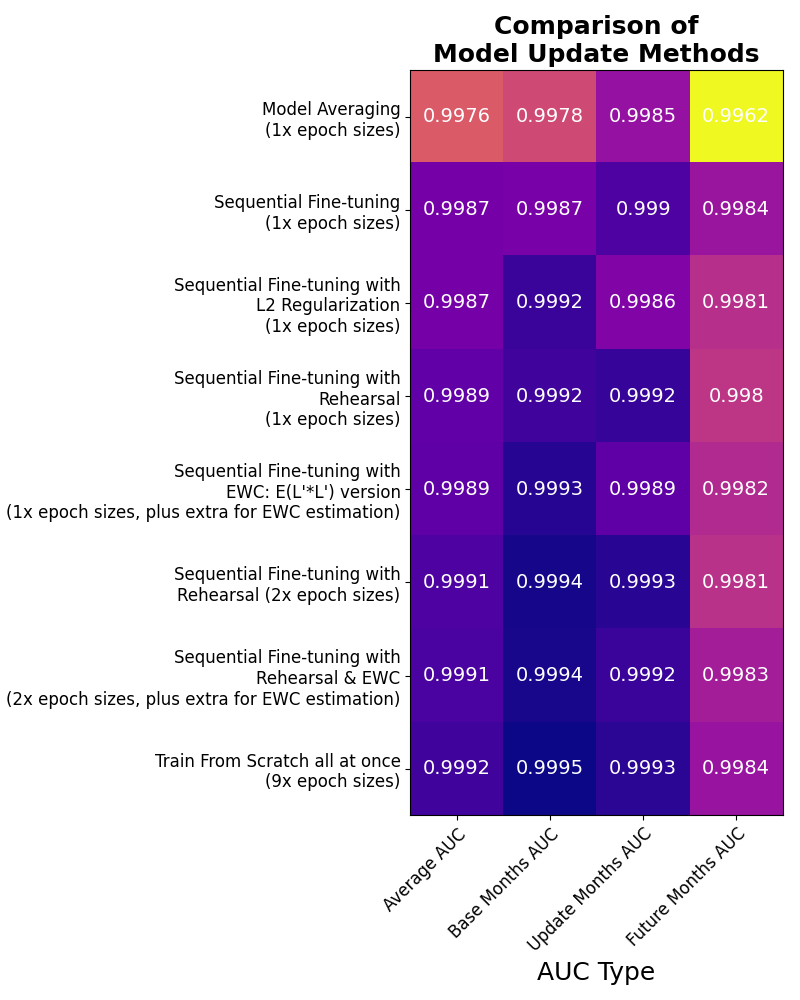}
    \caption{Various forms of model updates compared. For each method category (e.g. L2 regularization), the version with the best validation accuracy was selected (e.g. L2 with a certain Lambda value).}
\end{figure}

Our experiments concluded that if you do have access to past data, rehearsal is an excellent way to minimize catastrophic forgetting. Furthermore, if you're willing to increase model fine-tuning time, it is much more effective (and easy to implement). It also can be combined with regularization approaches, like EWC (which resulted in the best results overall).

\begin{figure}[H]
    \centering
    \includegraphics[width=.75\textwidth]{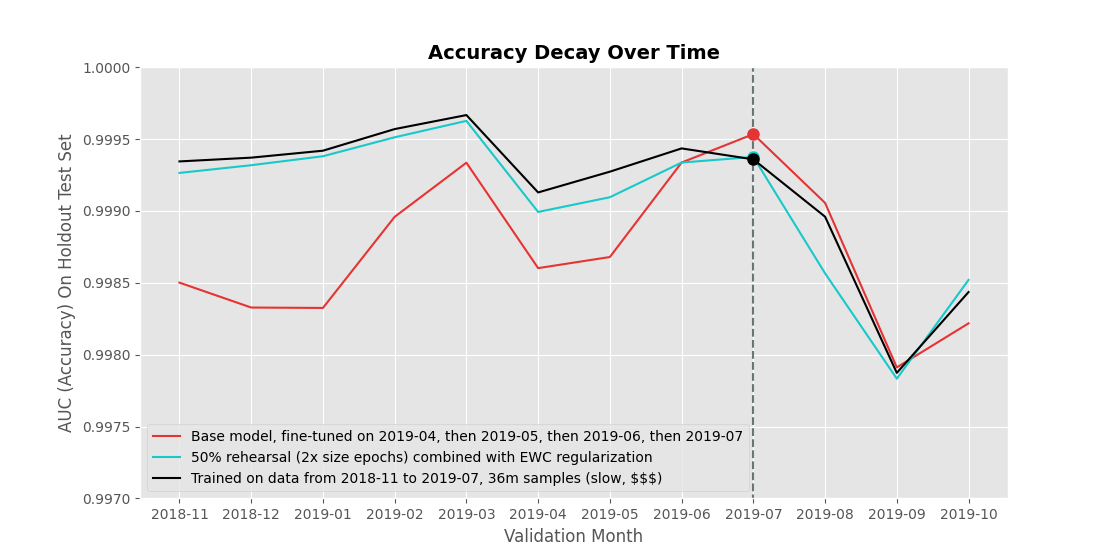}
    \caption{When data rehearsal is combined with EWC (aqua), the sequentially fine-tuned model very nearly matches the comparison model that is trained on all data all at once (black), and does much better than the baseline of sequential fine-tuning (red). Note that this version of data rehearsal did not fix epoch sizes, and so fine-tuning time is approximately twice that of normal (red's time), plus extra to estimate EWC damages (cost depends on how many samples you chose to estimate this with).} 
    \label{fig:datarehearsal4}
\end{figure}

\section*{Acknowledgements}
Thanks to Sophos for supporting this research.

\printbibliography

\end{document}